\documentclass{article}

\usepackage{PRIMEarxiv}

\usepackage[utf8]{inputenc} % allow utf-8 input
\usepackage[T1]{fontenc}    % use 8-bit T1 fonts
\usepackage{hyperref}       % hyperlinks
\usepackage{url}            % simple URL typesetting
\usepackage{booktabs}       % professional-quality tables
\usepackage{amsfonts}       % blackboard math symbols
\usepackage{nicefrac}       % compact symbols for 1/2, etc.
\usepackage{microtype}      % microtypography
\usepackage{lipsum}
\usepackage{fancyhdr}       % header
\usepackage{graphicx}       % graphics
\graphicspath{{images/}}     % organize your images and other figures under media/ folder

\usepackage[dvipsnames]{xcolor}
\usepackage{amsmath, amssymb}
\usepackage{notoccite}
\usepackage{float}
\usepackage{booktabs}
\usepackage{makecell}
\usepackage{algorithm}
\usepackage{algpseudocode}
\usepackage{algorithmicx}
\usepackage[caption=false,font=footnotesize]{subfig}
\usepackage{tikz}
\usetikzlibrary{positioning, shapes.geometric, shapes.multipart, shapes.symbols, calc}

\usepackage{orcidlink}

\title{EmbeddGAN: A Novel GAN Framework Using an Embedding Network and Gini Distance Correlation
\thanks{This work has been submitted to the IEEE for possible publication.
Copyright may be transferred without notice, after which this version may no
longer be accessible.}
}

\author{
  MaTais Caldwell\thanks{Corresponding author.}~\orcidlink{0000-0003-1608-6729} \\
  Department of Computer and Information Science \\
  University of Mississippi \\
  Oxford, MS 38677 USA \\
  \texttt{mkcaldwe@go.olemiss.edu} \\
  \And
  Yixin Chen~\orcidlink{0009-0009-5275-3665} \\
  Department of Computer and Information Science \\
  University of Mississippi \\
  Oxford, MS 38677 USA \\
  \And
  Xin Dang~\orcidlink{0000-0002-4328-4417} \\
  Department of Mathematics \\
  University of Mississippi \\
  Oxford, MS 38677 USA \\
  \And
  Charles Walter~\orcidlink{0000-0003-3063-6977} \\
  Department of Computer and Information Science \\
  University of Mississippi \\
  Oxford, MS 38677 USA \\
}

\begin{document}
\maketitle

\begin{abstract}
	Generative Adversarial Networks (GANs) have demonstrated strong performance in
	generating high-quality synthetic data. However, they are limited by no formal
	guarantees regarding convergence and the effectiveness of the learning process.
	In practice, this leads to training instability, mode collapse, and sensitivity
	to hyperparameters. To address this, we propose EmbeddGAN, a novel adversarial
	training framework based on a \emph{dependence-based objective}. Instead of
	relying on a discriminator that classifies samples as real or fake, EmbeddGAN
	introduces an embedding network that learns a representation in which
	statistical dependence between samples and their real/fake labels is maximized,
	while the generator is trained to minimize this dependence. This objective is
	implemented using the Gini distance correlation (gCor), which equals zero if
	and only if the embeddings are statistically independent of the real/fake
	label. Minimizing this objective therefore encourages real and generated
	samples to become statistically indistinguishable in the learned embedding
	space. The embedding network projects both real and generated data into a
	shared low-dimensional space, where distributional discrepancies can be
	measured directly through pairwise distances. We adopt a minimax training
	strategy: the embedding network maximizes the Gini distance correlation
	(maximizing dependence), while the generator minimizes it (minimizing
	dependence). Experiments on the MNIST, CIFAR-10, and CelebA datasets
	demonstrate that EmbeddGAN achieves competitive performance relative to
	established baselines while exhibiting notably stable training dynamics on the
	evaluated datasets.
\end{abstract}

% keywords can be removed

\section{Introduction}
\label{sec:introduction}
Generative Adversarial Networks (GANs) have established themselves as an
important framework for generative modeling since their
inception~\cite{goodfellowGenerativeAdversarialNetworks2014a}. By using an
adversarial framework involving a generator and discriminator, GANs are capable
of producing high-quality synthetic data that closely resembles real-world data
distributions. Their applications span across diverse domains, including image
synthesis~\cite{radfordUnsupervisedRepresentationLearning2016a,karrasStyleBasedGeneratorArchitecture2019a},
data augmentation~\cite{salimansImprovedTechniquesTraining2016a},
privacy-preserving data generation~\cite{shateriPreservingPrivacyGANs2023a},
and secure content creation~\cite{saxenaGenerativeAdversarialNetworks2022a}.

In addition to their ability to produce synthetic data, GANs are very
adaptable. Variants like conditional
GANs~\cite{mirzaConditionalGenerativeAdversarial2014a} allow for controlled
generation based on specific input conditions, while models like
StyleGAN~\cite{karrasStyleBasedGeneratorArchitecture2019a} enable fine-grained
manipulation of generated outputs. This flexibility has opened new
possibilities in creative industries, such as text-driven image
inpainting~\cite{leeDAFTGANDualAffine2024a}, natural image
generation~\cite{shahamSinGANLearningGenerative2019a}, and style
transfer~\cite{zhuUnpairedImagetoImageTranslation2017}. Furthermore, GANs have
been leveraged for privacy-preserving process-data generation, as demonstrated
by ProcessGAN~\cite{liProcessGANGeneratingPrivacyPreserving2024a}.

Despite their transformative potential, GANs are plagued by several persistent
challenges. GANs are prone to training
instability~\cite{arjovskyPrincipledMethodsTraining2017a}, as the adversarial
dynamic between the generator and discriminator often leads to oscillations and
non-convergence. This instability is compounded by mode
collapse~\cite{arjovskyPrincipledMethodsTraining2017a}, where the generator
produces limited or repetitive outputs, failing to capture the full diversity
of the data distribution.

Beyond these practical issues, the theoretical foundations of GAN training
remain limited. The standard GAN objective optimizes the Jensen-Shannon
divergence, which saturates and yields vanishing gradients when the real and
generated distributions have disjoint
support~\cite{arjovskyPrincipledMethodsTraining2017a}, providing no formal
guarantee that training converges to a meaningful equilibrium. While
Wasserstein GAN~\cite{arjovskyWassersteinGenerativeAdversarial2017} replaces
the Jensen-Shannon divergence with the earth movers distance, its guarantees
depend on exact enforcement of Lipschitz constraints. Formally grounded GAN
training objectives remain scarce, motivating the exploration of alternative
loss functions with well-defined statistical properties. Regardless of the
choice of divergence measure, the binary classification formulation itself is a
source of instability. A dependence-based objective sidesteps this entirely by
replacing the real/fake classifier with a principled measure of statistical
alignment between distributions

To explore alternative approaches to GAN training while preserving the
adversarial dynamic, this paper proposes \emph{EmbeddGAN}, a novel method
where, instead of training a generator against a traditional discriminator, we
introduce an \emph{embedding} network that projects data into a
lower-dimensional embedding space. This embedding network serves as a
\emph{feature extractor} that maps data to compact representations in a
low-dimensional embedding space. These embeddings are then used to compute the
\emph{Gini distance correlation
	loss}~\cite{zhangEstimatingFeatureLabelDependence2021}. This loss quantifies
statistical dependence between data and their labels (real or generated) by
contrasting within-group and between-group distances in the shared embedding
space. Gini distance correlation loss provides a formal guarantee of
distributional alignment: when the loss is minimized, the real and generated
data distributions become statistically indistinguishable in the embedding
space.

We use stochastic gradient descent to perform minimax training: unlike
traditional GAN frameworks, the embedding network is updated to maximize the
loss (maximizing dependence), while the generator is updated to minimize the
loss, encouraging adversarial learning in the embedding space. This process
ensures that the generator is continually challenged to produce high-quality
synthetic data. This approach represents a departure from classical
discriminator-based-training, exploring an alternative adversarial learning
paradigm.

Lastly, EmbeddGAN is architecturally agnostic: the Gini distance correlation
loss is completely decoupled from the specific architecture of the embedding
network, allowing for flexibility in choosing different convolutional
architectures or projection strategies. This means that our method can be
easily adapted to various GAN architectures, making it a versatile tool for
generative modeling across different domains and applications.

\noindent The contributions of this paper are as follows:
\begin{itemize}
	\item We propose a novel training strategy that replaces the traditional GAN
	      discriminator with an embedding network trained using the Gini distance
	      correlation loss.

	\item We introduce an adversarial feature space learning mechanism via minimizing the
	      negative Gini distance correlation, creating a novel training dynamic that
	      differs from traditional GAN approaches.

	\item We conduct a controlled empirical evaluation on MNIST, CIFAR-10, and CelebA,
	      demonstrating competitive performance against DCGAN, WGAN-GP, and SN-GAN while
	      exhibiting notably stable training dynamics.

	\item We provide a systematic ablation analysis, including Fourier spectrum
	      diagnostics, embedding dimension sensitivity, and update ratio effects. Lastly,
	      we offer practical insight into the design requirements of dependence-based
	      adversarial training.
\end{itemize}

The remainder of this paper is structured as follows:
Section~\ref{sec:background} reviews the classical GAN formulation and the Gini
distance correlation. In Section~\ref{sec:methods}, we present our model
architecture. Section~\ref{sec:experimental_setup} describes our experimental
setup, and Section~\ref{sec:results} presents our results. Finally,
Section~\ref{sec:conclusion} concludes the paper and discusses future work.

\section{Background}
\label{sec:background}

In this section we provide a brief overview of the traditional GAN formulation
and discuss some of its issues. We end with a discussion of the Gini distance
correlation and its properties.

\subsection{Generative Adversarial Networks}
\begin{figure}[ht]
	\centering
	\scalebox{0.6}{
		\begin{tikzpicture}[
				node distance=1cm and 1.4cm,
				every node/.style={font=\bfseries, align=center},
				block/.style={draw, rounded corners, minimum width=2.2cm,
						minimum height=1.3cm, text width=2.2cm, align=center},
				arrow/.style={->, thick},
				revarrow/.style={->, thick, dashed}
			]

			% Nodes
			\node[block, fill=gray!10] (gen) {Generator \\ $G$};
			\node[above=0.5cm of gen] (zlabel) {$z \sim \mathcal{N}(0,1)$};
			\draw[arrow] (zlabel) -- (gen);

			\node[block, fill=gray!10, right=1.0cm of gen] (x_fake) {Fake Image \\ $x_{\text{fake}}$};
			\draw[arrow] (gen) -- (x_fake);

			\node[block, fill=gray!10, above=0.5cm of x_fake] (x_real) {Real Image \\ $x_{\text{real}}$};

			\node[block, fill=gray!10, right=1.0cm of x_fake] (disc) {Discriminator \\ $D$};

			\node[block, fill=gray!10, right=1.0cm of disc] (prob) {Real / Fake \\ Probability};

			\node[draw, cloud, cloud ignores aspect, fill=gray!10, text width=1.7cm,
				align=center, minimum height=1cm, below=0.5cm of prob]
			(loss) {Binary Cross-Entropy \\ Loss (Min/Max)};

			% Arrows
			\draw[arrow] (x_fake) -- (disc);
			\draw[arrow] (x_real) -- (disc);
			\draw[arrow] (disc) -- (prob);
			\draw[arrow] (prob) -- (loss);

			% Backpropagation arrows
			\draw[revarrow, color=black!65](loss.west) to[out=180, in=-90]
			node[midway, left, xshift=-0.2cm] {Discriminator Update}
			(disc.south);
			\draw[revarrow, color=black!65](loss.west) to[out=180, in=-90]
			node[midway, left, xshift=-1.8cm] {Generator Update}
			(gen.south);

			% Labels
			\node[above=0.1cm of prob] {\footnotesize Output Space};
			\node[above=0.9cm of gen] {\footnotesize Latent Vector};

		\end{tikzpicture}
	}
	\caption{Overview of the traditional GAN architecture.}
	\label{fig:gan_architecture}
\end{figure}
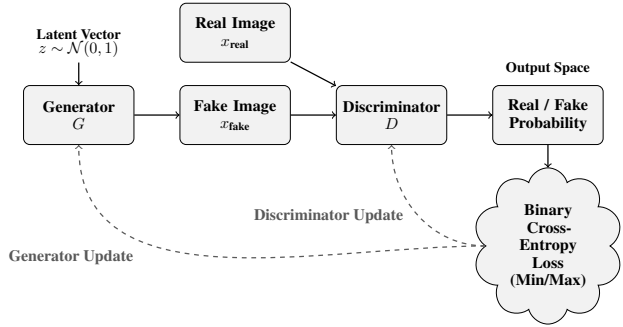

A basic GAN consists of two neural networks, a generator and a discriminator,
that play an adversarial game. The generator $G$ takes in a random noise $z$,
typically drawn from a Gaussian, and tries to produce samples that are
indistinguishable from real data. The discriminator $D$ takes as input both
real and generated samples, and tries to correctly classify them as either real
or fake. The generator continually tries to fool the discriminator by
generating samples that are increasingly similar to real data. An overview of a
generalized traditional GAN architecture is shown in
Fig.~\ref{fig:gan_architecture}.

Formally, the game between the generator and discriminator can be expressed as
a minimax optimization problem:

\begin{equation}
	\min_G \max_D V(D, G)  = \mathbb{E}_{x \sim p_{\text{data}}(x)}[\log D(x)]
	+ \mathbb{E}_{z \sim p_z(z)}[\log(1 - D(G(z)))]
\end{equation}

where $p_{\text{data}}$ is the distribution of real data, and $p_z$ is the
distribution of random noise.

Typically, the discriminator becomes much better at distinguishing real samples
from fake ones, leading to vanishing gradients for the generator. This can
cause the generator to produce low-quality samples or even experience mode
collapse. To mitigate these issues, several alternatives to the traditional GAN
objective have been proposed. Wasserstein GANs replace the Jenson-Shannon
divergence with the Earth Mover's distance to improve training
dynamics~\cite{arjovskyPrincipledMethodsTraining2017a}. Other approaches, such
as MMD-GANs~\cite{li2017mmd}, leverage kernel-based distances like Maximum Mean
Discrepancy to evaluate differences between distributions. These metrics offer
stability and theoretical grounding but often require careful tuning of
hyperparameters and kernel choices. Other papers have explored using learned
feature spaces rather than raw pixel space to compute similarity between real
and generated
data~\cite{donahue2016adversarial,dumoulinAdversariallyLearnedInference2017}.
In contrast with MMD-GANs, the embedding space in EmbeddGAN is adversarially
learned rather than fixed.

Building on feature-based approaches, projections have emerged as an effective
tool for dimensionality reduction in generative
modeling~\cite{blumRandomProjectionMargins2006,rahimiRandomFeaturesLargeScale2007}.
Several works improve training stability and sample quality using multiple
discriminators or ensemble strategies. For example, Neyshabur et
al.~\cite{neyshaburStabilizingGANTraining2018a} train a single generator
against multiple discriminators, each operating on a different random
low-dimensional projection. Other multi-discriminator methods assign distinct
discriminators to different aspects of the data
distribution~\cite{bhattacharyaGenerativeAdversarialSpeaker2019a}. Improved GAN
training techniques~\cite{salimansImprovedTechniquesTraining2016a} further
highlight the role of feature-based evaluation and alternative loss functions.
Our work extends these insights by combining learned convolutional features
with learned projections.
\subsection{Gini Distance Correlation}
\label{sec:gini_distance_correlation}

We adopt the \emph{Gini distance correlation} proposed by Zhang et
al.~\cite{zhangEstimatingFeatureLabelDependence2021} as our loss function. This
statistic quantifies the dependence between a continuous variable and a
categorical label using Gini mean differences. It equals zero \emph{if and only
	if} the variables are statistically independent, making it well-suited for
distributional alignment tasks in generative modeling.

Let $X \in \mathbb{R}^n$ be a random variable from distribution $F$, and let $Y
	\in \{0, 1\}$ be a categorical variable with conditional distributions $F_0$
and $F_1$ over $X$. Define $p_k = \mathbb{P}(Y = k)$, and let $X_k \sim F_k$,
with $X'$ and $X_k'$ denoting independent copies of $X$ and $X_k$,
respectively. The \emph{Gini distance covariance} is defined as:

\begin{equation}
	\text{gCov}(X, Y)  = \sum_{k=0}^{1} p_k [ 2 \, \mathbb{E}\|X_k - X\|
	- \mathbb{E}\|X_k - X_k'\| - \mathbb{E}\|X - X'\|]
\end{equation}

The corresponding \emph{Gini distance correlation} is the normalized form:
\begin{align}
	\rho_g(X, Y) = \frac{\text{gCov}(X, Y)}{\mathbb{E} \|X - X'\|}
\end{align}
This satisfies $\rho_g(X, Y) \in [0, 1]$, and equals zero \emph{if and only if} $X
	\perp Y$.

We estimate $\rho_g$ from batches of embeddings. Let $A = \{a_i\}_{i=1}^m$, $B
	= \{b_j\}_{j=1}^m$, and $X = A \cup B = \{x_1, \dots, x_{2m}\}$, where $Y = 1$
for real and $Y = 0$ for generated samples. The empirical Gini distance
covariance and normalizing factor are:

\begin{equation}
	\text{gCov}(A, B)  = \frac{1}{m^2} \sum_{i, j} \left\|a_i - b_j\right\|
	- \frac{1}{2m(m-1)} \sum_{i \ne j} \left[\|a_i - a_j\| + \|b_i - b_j\|\right]
	\label{eq:gmd_within}
\end{equation}

\begin{equation}
	\Delta(X)          = \frac{1}{{(2m)}^2} \sum_{i, j} \|x_i - x_j\|
	\label{eq:delta}
\end{equation}
The empirical Gini distance correlation is:
\begin{align}
	\rho_g(A, B) = \frac{\text{gCov}(A, B)}{\Delta(X)} \label{eq:gcor}
\end{align}

This normalized formulation ensures that $\rho_g(A, B) \in [0, 1]$, and equals
zero when the real and generated distributions are statistically
indistinguishable in the embedding space.

Minimizing $\rho_g$ therefore encourages the combined variable $X = A \cup B$
to become statistically independent of the binary label $Y \in \{0, 1\}$. In
other words, when the distributions of real and generated samples align, the
label becomes uninformative with respect to the embeddings, resulting in
$\rho_g = 0$. This makes the Gini distance correlation a natural objective for
distributional alignment in generative modeling. Compared to adversarial
losses, this approach offers a symmetric, differentiable objective with a clear
probabilistic interpretation. In the univariate case, the intra-group terms
reduce to the Gini Mean Difference (GMD).

This loss is core to our framework. In the next section, we describe our model
architecture, which leverages this loss to create a novel adversarial training
dynamic.
\section{Methods}
\label{sec:methods}

\subsection{Model Architecture}
At the heart of EmbeddGAN is the Gini distance correlation, which quantifies
the statistical dependence between data and their labels. Given labels of real
or generated data, a Gini distance correlation of zero implies statistical
independence, i.e., the real and generated data come from the same
distribution. Unlike classical discriminators trained to classify samples as
real or fake, our embedding network aims to identify a low-dimensional space
that maximizes this dependence. The generator attempts to minimize it. This
adversarial objective is implemented via stochastic gradient descent, both
networks having opposing goals: the embedding network increases dependence by
maximizing the Gini distance correlation, and the generator decreases its loss
by minimizing it. This leads to an adversarial learning process grounded not in
classification accuracy, but in implicitly reducing the divergence between
generated and real data distributions within a shared embedding space. Our
adversarial training dynamic is formalized by the following minimax objective:
\begin{align}
	\min_G \max_E \; \rho_g\left( E(G(z)), E(x_{\text{real}}) \right)
\end{align}
where $\rho_g(\cdot, \cdot)$ denotes the empirical Gini distance correlation between
the embedded fake and real samples.

The embedding network defines the learned feature space in which statistical
distances are computed. Rather than relying on fixed similarity kernels as in
MMD-GANs~\cite{li2017mmd}, our approach uses a learnable convolutional
architecture followed by a linear projection to embed both real and generated
images into a shared low-dimensional space. Across datasets, the embedding
network follows a consistent structural pattern mirroring the DCGAN
discriminator~\cite{radfordUnsupervisedRepresentationLearning2016a}: stacked
stride-2 \texttt{Conv2d} layers with LeakyReLU activations and dropout,
followed by a flattening step and a final linear projection into an
$n$-dimensional embedding space. We set $n = 100$ for our experiments, but we
find any value of $n\geq3$ to be sufficient. Spectral
normalization~\cite{miyatoSpectralNormalizationGenerative2018} is applied to
all convolutional and linear layers to stabilize training by constraining the
Lipschitz constant of the embedding network. While exact dimensions and input
channels vary between datasets (e.g., grayscale vs. RGB), the design principle
remains the same: learn task-relevant representations while retaining
sufficient flexibility for distributional alignment.

The generator follows the DCGAN
architecture~\cite{radfordUnsupervisedRepresentationLearning2016a}, consisting
of strided transpose convolutional layers with batch normalization, with
LeakyReLU activations used in place of ReLU across all layers to ensure
consistency across model components. All convolutional and linear weights are
initialized using a Gaussian distribution with mean $0$ and standard deviation
of $0.02$, a choice known to promote stable GAN
training~\cite{radfordUnsupervisedRepresentationLearning2016a}. This
architecture enables the model to compare real and generated data meaningfully
through Gini distance correlation. An overview of our generator architectures
can be seen in Table~\ref{tab:gen_arch}.

\begin{table}[t]

	\centering

	\caption{Generator architectures. BN = batch normalization;
		LReLU = LeakyReLU(0.2); $k$ = kernel size; $s$ = stride. LeakyReLU is used in
		place of ReLU across all layers to ensure consistency across model components.}

	\label{tab:gen_arch}

	\renewcommand{\arraystretch}{1.25}

	% ---- MNIST ----
	\textbf{MNIST}\\[4pt]
	\begin{tabular}{@{}lc@{}}
		\toprule
		\textbf{Layer}                                           & \textbf{Output Shape}     \\
		\midrule
		Input $z \sim \mathcal{N}(\mathbf{0}, \mathbf{I}_{100})$ & $100$                     \\
		Linear, BN, LReLU, Reshape                               & $256 \times 7 \times 7$   \\
		ConvT $256\!\to\!128$, $k{=}5$, $s{=}1$, BN, LReLU       & $128 \times 7 \times 7$   \\
		ConvT $128\!\to\!64$,  $k{=}5$, $s{=}2$, BN, LReLU       & $64  \times 14 \times 14$ \\
		ConvT $64\!\to\!1$,    $k{=}5$, $s{=}2$, Tanh            & $1   \times 28 \times 28$ \\
		\bottomrule
	\end{tabular}

	\medskip

	% ---- CIFAR-10 ----
	\textbf{CIFAR-10}\\[4pt]
	\begin{tabular}{@{}lc@{}}
		\toprule
		\textbf{Layer}                                           & \textbf{Output Shape}     \\
		\midrule
		Input $z \sim \mathcal{N}(\mathbf{0}, \mathbf{I}_{100})$ & $100$                     \\
		Linear, BN, LReLU, Reshape                               & $512 \times 4  \times 4$  \\
		ConvT $512\!\to\!256$, $k{=}4$, $s{=}2$, BN, LReLU       & $256 \times 8  \times 8$  \\
		ConvT $256\!\to\!128$, $k{=}4$, $s{=}2$, BN, LReLU       & $128 \times 16 \times 16$ \\
		ConvT $128\!\to\!3$,   $k{=}4$, $s{=}2$, Tanh            & $3   \times 32 \times 32$ \\
		\bottomrule
	\end{tabular}

	\medskip

	% ---- CelebA ----
	\textbf{CelebA}\\[4pt]
	\begin{tabular}{@{}lc@{}}
		\toprule
		\textbf{Layer}                                           & \textbf{Output Shape}     \\
		\midrule
		Input $z \sim \mathcal{N}(\mathbf{0}, \mathbf{I}_{100})$ & $100$                     \\
		Linear, BN, LReLU, Reshape                               & $512 \times 4  \times 4$  \\
		ConvT $512\!\to\!256$, $k{=}5$, $s{=}2$, BN, LReLU       & $256 \times 8  \times 8$  \\
		ConvT $256\!\to\!128$, $k{=}5$, $s{=}2$, BN, LReLU       & $128 \times 16 \times 16$ \\
		ConvT $128\!\to\!64$,  $k{=}5$, $s{=}2$, BN, LReLU       & $64  \times 32 \times 32$ \\
		ConvT $64\!\to\!3$,    $k{=}5$, $s{=}2$, Tanh            & $3   \times 64 \times 64$ \\
		\bottomrule
	\end{tabular}
\end{table}

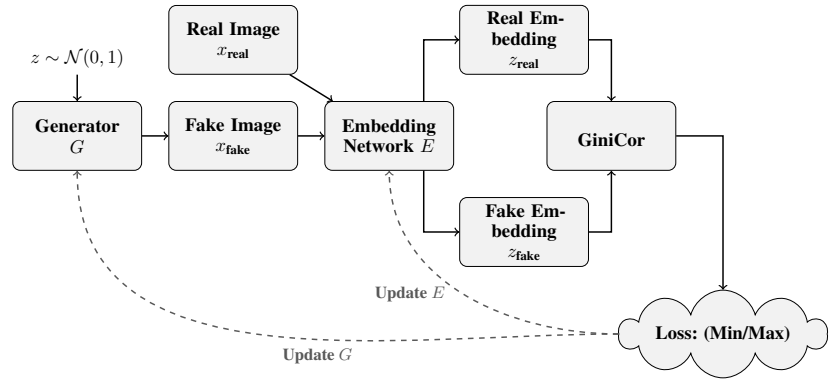
\begin{figure}[ht!]
	\centering
	\scalebox{0.7}{
		\begin{tikzpicture}[
				node distance=1.2cm and 1.4cm,
				every node/.style={font=\bfseries, align=center},
				block/.style={draw, rounded corners, minimum width=2.2cm, minimum height=1.3cm, text width=2.2cm, align=center},
				arrow/.style={->, thick},
				revarrow/.style={->, thick, dashed}
			]

			% Nodes
			\node[block, fill=gray!10] (gen) {Generator \\ $G$};
			\node[above=0.5cm of gen] (zlabel) {$z \sim \mathcal{N}(0,1)$};
			\draw[arrow] (zlabel) -- (gen);

			\node[block, fill=gray!10, right=0.5cm of gen] (x_fake) {Fake Image \\ $x_{\text{fake}}$};
			\draw[arrow] (gen) -- (x_fake);

			\node[block, fill=gray!10, above=0.5cm of x_fake] (x_real) {Real Image \\ $x_{\text{real}}$};

			\node[block, fill=gray!10, right=0.5cm of x_fake] (embed) {Embedding \\ Network $E$};

			% Embedding output nodes
			\node[block, fill=gray!10, above right=0.5cm and 0.1cm of embed] (z_real) {Real Embedding \\ $z_{\text{real}}$};
			\node[block, fill=gray!10, below right=0.5cm and 0.1cm of embed] (z_fake) {Fake Embedding \\ $z_{\text{fake}}$};

			% Gini Correlation node
			\node[block, fill=gray!10, right=1.75cm of embed] (gini) {\\GiniCor};

			% Loss node moved downward
			\node[draw, cloud, cloud ignores aspect, fill=gray!10,
				below right=2.5cm and -0.5cm of gini, minimum width=2.2cm, minimum height=1.2cm, align=center] (loss) {Loss: (Min/Max)};

			% Arrows from real/fake to embed
			\draw[arrow] (x_fake) -- (embed);
			\draw[arrow] (x_real) -- (embed);

			% Arrows from embed to embeddings
			\draw[arrow] ([xshift=18mm] embed) |- (z_real);
			\draw[arrow] ([xshift=18mm] embed) |- (z_fake);

			% Arrows from embeddings to GiniCor
			\draw[arrow] (z_real) -| (gini);
			\draw[arrow] (z_fake) -| (gini);

			% Arrow from GiniCor to Loss (downward)
			\draw[arrow] (gini) -| (loss);

			% Backprop arrows from Loss
			\draw[revarrow, color=black!65] (loss.west) to[out=180, in=-90]
			node[midway, left, xshift=-0.2cm]
			{\footnotesize \textbf{Update $E$}} (embed.south);

			\draw[revarrow, color=black!65] (loss.west) to[out=180, in=-90]
			node[midway, right, xshift=0.2cm, yshift=-0.4cm]
			{\footnotesize \textbf{Update $G$}} (gen.south);

		\end{tikzpicture}
	}
	\caption{Overview of the EmbeddGAN architecture. Both real and generated images are projected into a shared embedding space
		by the embedding network $E$. The Gini distance correlation compares the distributions of
		real and fake embeddings.}
	\label{fig:embeddgan_architecture_column_fit}
\end{figure}

It is important to note that EmbeddGAN is architecture-agnostic. The Gini
distance correlation loss operates entirely in a learned embedding space. The
loss has no structural coupling with the architectures that produce these
embeddings. The generator can be any differentiable mapping from latent vector
to image (convolutional, residual, transformer-based, etc.), and the embedding
network can be any differentiable encoder. In this work, DCGAN-style generators
and convolutional encoders are used across all datasets to ensure a fair
comparison with the DCGAN, WGAN-GP, and SN-GAN baselines; however, EmbeddGAN
should be understood as a \emph{general training objective and optimization
	protocol} rather than a fixed architecture, and is applicable to more
expressive generator designs such as residual or attention-based networks. An
overview of the full architecture is shown in
Fig.~\ref{fig:embeddgan_architecture_column_fit}.

\subsection{Regularization of the Embedding Network}

\label{sec:regularization}

An embedding network trained adversarially without constraints is susceptible
to a shortcut exploitation. Without bounds on the embedding vectors, the
network can separate real from generated samples by exploiting the magnitude of
the embedding vectors rather than their distributional structure, a degenerate
solution that does not require the generator to produce realistic samples. We
therefore apply \emph{L2 normalization} to all embedding outputs, projecting
them onto the unit hypersphere and preventing any dimension from dominating the
representation. However, constraining the magnitude alone is insufficient: once
the magnitude shortcut is closed, the embedding network can exploit
high-contrast regions in the input as an alternative shortcut. We apply
\emph{R1 gradient regularization}~\cite{meschederWhichTrainingMethods2018} to
the embedding network, penalizing the squared norm of gradients with respect to
real samples. This imposes a smoothness constraint on the embedding surface,
removing the gradient incentive to latch onto high-frequency or high-contrast
features. Together, L2 normalization and R1 regularization target distinct
failure modes and are jointly necessary.
% The ablation study in
% Section~\ref{sec:ablation_study} provides quantitative evidence that neither
% component alone is sufficient.

\section{Experimental Setup}
\label{sec:experimental_setup}

\subsection{Datasets}
\label{sec:datasets}

We choose to show viability of EmbeddGAN on three common datasets, the
MNIST~\cite{lidengMNISTDatabaseHandwritten2012},
CIFAR-10~\cite{krizhevskyLearningMultipleLayers2009}, and
CelebA~\cite{liuDeepLearningFace2014} datasets. Each dataset was chosen to show
the viability of EmbeddGAN on different types of image data and at different
scales. The MNIST dataset contains $10$ digit classes, and $28 \times 28$
grayscale images. CIFAR-10 consists of $32 \times 32$ color images with $10$
distinct classes. CelebA faces contains over $200,000$ celebrity images with
$40$ attribute annotations. For our experiments, we use the aligned and cropped
version of CelebA, which consists of $178 \times 218$ RGB images. We resize
these images to $64 \times 64$ for training.

\subsection{Models for Comparison}

\label{sec:models_for_comparison}

To evaluate the performance of EmbeddGAN, we compare it to three well-known GAN
variants, DCGAN~\cite{radfordUnsupervisedRepresentationLearning2016a},
Wasserstein GAN with gradient penalty
(WGAN-GP)~\cite{gulrajaniImprovedTrainingWasserstein2017a}, and Spectral Norm
GAN (SN-GAN)~\cite{miyatoSpectralNormalizationGenerative2018}. Each baseline
was selected deliberately, not merely as a point of comparison, but because
each shares a structural or conceptual relationship with EmbeddGAN. DCGAN
serves as a foundational baseline. Both EmbeddGAN's embedding network and
DCGAN's discriminator share the same convolutional backbone: stacked stride-2
\texttt{Conv2d} layers with LeakyReLU activations, followed by a final linear
projection. This architectural similarity ensures that performance differences
between the two models can be attributed to the training objective rather than
to architectural capacity.

WGAN-GP is included because its training objective pursues a similar goal to
the Gini distance correlation: implicit distributional alignment between real
and generated data. Where WGAN-GP measures this alignment via the Wasserstein
distance enforced through a gradient penalty, EmbeddGAN measures it through
statistical dependence in a learned embedding space. Comparing the two allows
us to examine whether \emph{gCor} offers a competitive alternative to
Wasserstein-based objectives on these benchmarks.

SN-GAN is chosen because EmbeddGAN's embedding network itself employs spectral
normalization on its convolutional and linear layers. Since spectral
normalization is the defining characteristic of SN-GAN's discriminator,
including it as a baseline allows for a direct comparison between a model that
applies spectral normalization in a classical adversarial setting and one that
applies it within an embedding-based framework.

To ensure a fair comparison, all three baselines share the same generator and
discriminator architectures, differing only at the layer level: DCGAN includes
batch normalization in the discriminator, while WGAN-GP and SN-GAN do not.
SN-GAN replaces batch normalization with spectral normalization on all
discriminator layers.

\subsection{Training Algorithm and Hyperparameters}
\begin{algorithm}
	\caption{EmbeddGAN Training}
	\label{alg:embeddgan}
	\footnotesize
	\begin{algorithmic}[1]
		\Require Generator $G_{\theta_G}$, embedding network $E_{\theta_E}$ (L2-normalised output)
		\Require Batch size $m$, latent dimension $z_d$, embedding dimension $n$
		\Require Learning rate $\eta$, Adam $(\beta_1, \beta_2)$
		\Require Embedding steps $k$, R1 coefficient $\gamma$
		\For{epoch $t = 1, \ldots, T$}
		\For{each mini-batch $\mathbf{x}^r \sim p_{\mathrm{data}}$, $|\mathbf{x}^r| = m$}

		\State \textbf{// Phase 1 --- update embedding network ($k$ steps)}
		\State Sample $\mathbf{z} \sim \mathcal{N}(\mathbf{0}, \mathbf{I})^{m \times z_d}$
		\State $\mathbf{x}^f \leftarrow G_{\theta_G}(\mathbf{z})$ \Comment{freeze $\theta_G$; no gradient}
		\For{$s = 1, \ldots, k$}
		\State $\mathbf{e}^r \leftarrow E_{\theta_E}(\mathbf{x}^r),\quad
			\mathbf{e}^f \leftarrow E_{\theta_E}(\mathbf{x}^f)$
		\State Compute ${\mathrm{\rho_g}}(\mathbf{e}^r, \mathbf{e}^f)$
		via Eqs.~\eqref{eq:gmd_within}--\eqref{eq:gcor}
		\State $\mathcal{R}_1 \leftarrow \dfrac{1}{m}
			\displaystyle\sum_{n=1}^{m}
			\!\left\|
			\nabla_{\mathbf{x}^r_n}
			\textstyle\sum_{\ell} E_{\theta_E}(\mathbf{x}^r_n)_\ell
			\right\|^2$
		\State $\mathcal{L}_E \leftarrow -{\mathrm{\rho_g}} +
			\dfrac{\gamma}{2}\,\mathcal{R}_1$
		\State $\theta_E \leftarrow \mathrm{Adam}\!\left(\theta_E,\,
			\nabla_{\theta_E}\mathcal{L}_E,\, \eta,\, \beta_1,\, \beta_2\right)$
		\EndFor

		\State \textbf{// Phase 2 --- update generator (1 step)}
		\State Sample $\mathbf{z} \sim \mathcal{N}(\mathbf{0}, \mathbf{I})^{m \times z_d}$
		\State $\mathbf{x}^f \leftarrow G_{\theta_G}(\mathbf{z})$ \Comment{freeze $\theta_E$}
		\State $\mathbf{e}^r \leftarrow E_{\theta_E}(\mathbf{x}^r),\quad
			\mathbf{e}^f \leftarrow E_{\theta_E}(\mathbf{x}^f)$
		\State $\mathcal{L}_G \leftarrow {\mathrm{\rho_g}}(\mathbf{e}^r, \mathbf{e}^f)$
		\State $\theta_G \leftarrow \mathrm{Adam}\!\left(\theta_G,\,
			\nabla_{\theta_G}\mathcal{L}_G,\, \eta,\, \beta_1,\, \beta_2\right)$

		\EndFor
		\EndFor
	\end{algorithmic}
\end{algorithm}

EmbeddGAN alternates between two decoupled optimization phases per mini-batch,
as summarized in Algorithm~\ref{alg:embeddgan}. Rather than the adversarial
minimax game of standard GANs, the embedding network and generator pursue
complementary objectives with respect to the Gini distance correlation
$\rho_g$: the embedding network is trained to \emph{maximize} $\rho_g$,
increasing dependence between features and their labels, while the generator is
trained to \emph{minimize} $\rho_g$, decreasing this dependence by aligning the
distributions of real and generated data in this embedding space. The embedding
network applies L2 normalization to its outputs and is trained with R1 gradient
regularization, as described in Section~\ref{sec:regularization}

All models were trained with the hyperparameters listed in
Table~\ref{tab:hyperparams}. The same values were used across all three
datasets except where noted. We set $k = 1$ (one embedding update per generator
update) following preliminary experiments that showed no consistent benefit
from higher values. For WGAN-GP, we set $k = 5$ following the original paper's
recommendation of multiple critic updates per generator update.

\begin{table}[h]
	\centering
	\caption{EmbeddGAN hyperparameters.}
	\label{tab:hyperparams}
	\begin{tabular}{lc}
		\toprule
		Hyperparameter                   & Value              \\
		\midrule
		Epochs $T$                       & 500                \\
		Latent dimension $z_d$           & 100                \\
		Embedding dimension $n$          & 100                \\
		Learning rate $\eta$             & $2 \times 10^{-4}$ \\
		Adam $\beta_1$                   & 0.5                \\
		Adam $\beta_2$                   & 0.999              \\
		Batch size $m$ (MNIST, CIFAR-10) & 256                \\
		Batch size $m$ (CelebA)          & 1024               \\
		Embedding steps $k$              & 1                  \\
		R1 coefficient $\gamma$          & 0.5                \\
		\bottomrule
	\end{tabular}
\end{table}
\subsection{Evaluation}
We evaluate the models using the Fr\'echet Inception Distance
(FID)~\cite{heuselGANsTrainedTwo2018a}, a widely used metric for assessing the
quality of generated images. FID measures the similarity between the
distribution of real images and the distribution of generated images by
comparing their statistics in the feature space of a pretrained Inception v3
network. Lower FID indicates greater similarity between the two distributions,
reflecting both higher visual quality and greater diversity in generated
samples. To compute the FID for each model, we use the \texttt{clean-fid}
library~\cite{parmarAliasedResizingSurprising2021}.

Across the three datasets, well-trained GAN baselines typically achieve FID
scores in the following ranges: on MNIST, scores below $\sim$20 are readily
achievable, though FID is a less reliable indicator on this dataset due to the
domain mismatch between MNIST and the ImageNet-pretrained Inception v3
network~\cite{shmelkovHowGoodMy2018}. On CIFAR-10, DCGAN, WGAN-GP, and SN-GAN
can achieve FIDs in the range of approximately 14--55, with well tuned models
typically falling between
21--40~\cite{miyatoSpectralNormalizationGenerative2018,heuselGANsTrainedTwo2018a,lucicAreGANsCreated2018}.
On CelebA 64$\times$64, DCGAN typically achieves FIDs of 13--65 depending on
training strategy, WGAN-GP achieves 3--30, and SN-GAN achieves 10--30 depending
on the
architecture~\cite{heuselGANsTrainedTwo2018a,zhangConsistencyRegularizationGenerative2019}.

We also compute the recall of the generated samples, which measures the
diversity of the generated data by quantifying the fraction of real data modes
that are covered by the generated
distribution~\cite{kynkaanniemiImprovedPrecisionRecall2019}. This is
complementary to FID, which captures both quality and diversity but does not
explicitly measure mode coverage. We use the \texttt{pdrc} library to compute
recall~\cite{naeemReliableFidelityDiversity2020}, with k=5 and 10,000 samples
per model, using 2048-d Inception-v3 pool features in place of VGG-16.

\subsection{Computing Resources}
All experiments were conducted on a workstation with an NVIDIA RTX A6000 GPU
with 49GB of VRAM. The operating system was Debian 12, with CUDA 13.2 for GPU
acceleration. All models were implemented using PyTorch 2.12.0.

\section{Results}
\label{sec:results}

\subsection{Quantitative Comparison}

\subsubsection{Fr\'echet Inception Distance Results}
\begin{table}[]
	\centering
	\caption{Fr\'echet Inception Distance on the MNIST, CIFAR-10, and CelebA dataset at 500 epochs.
		Scores are computed over a sample size of 50000.}
	\begin{tabular}{cccc}
		\toprule\toprule
		Model     & MNIST $\downarrow$ & CIFAR-10 $\downarrow$ & CelebA $\downarrow$ \\
		\midrule
		EmbeddGAN & $6.59$             & $50.89$               & $32.09$             \\
		DCGAN     & $6.63$             & $41.92$               & $37.30$             \\
		WGAN-GP   & $9.71$             & $47.66$               & $30.67$             \\
		SN-GAN    & $7.96$             & $60.87$               & $23.97$             \\
		\bottomrule
	\end{tabular}
	\label{tab:FID Scores}
\end{table}

Table~\ref{tab:FID Scores} reports FID scores for all four models at epoch 500
across three datasets. On MNIST, EmbeddGAN ($6.59$) and DCGAN ($6.63$) converge
to near-identical performance, with SN-GAN ($7.96$) and WGAN-GP ($9.71$)
trailing. For CIFAR-10, DCGAN leads with a score of $41.92$; WGAN-GP ($47.66$)
and EmbeddGAN ($50.89$) are competitive, while SN-GAN ($60.87$) underperforms.
And on the CelebA dataset, SN-GAN achieves the lowest FID ($23.97$), having
continued to improve beyond epoch 300; WGAN-GP ($30.67$) and EmbeddGAN
($32.09$) are competitive, while DCGAN ($37.30$) trails. Across all three
datasets, no single model dominates; EmbeddGAN achieves competitive performance
in each setting without specializing to any one dataset.

\subsubsection{Recall Results}

\begin{table}[]
	\centering
	\caption{Recall on the MNIST, CIFAR-10, and CelebA dataset at 500 epochs.
		Scores are computed over a sample size of 10000.}
	\begin{tabular}{cccc}
		\toprule\toprule
		Model     & MNIST $\uparrow$ & CIFAR-10 $\uparrow$ & CelebA $\uparrow$ \\
		\midrule
		EmbeddGAN & $0.854$          & $0.469$             & $0.352$           \\
		DCGAN     & $0.844$          & $0.472$             & $0.376$           \\
		WGAN-GP   & $0.824$          & $0.483$             & $0.495$           \\
		SN-GAN    & $0.819$          & $0.328$             & $0.471$           \\
		\bottomrule
	\end{tabular}
	\label{tab:Recall Scores}
\end{table}

Table~\ref{tab:Recall Scores} reports Recall scores for all four models at
epoch 500 across three datasets. On MNIST, EmbeddGAN achieves the highest
recall ($0.854$), with DCGAN ($0.844$), WGAN-GP ($0.824$), and SN-GAN ($0.819$)
trailing. For CIFAR-10, WGAN-GP leads with a recall of $0.483$; DCGAN ($0.472$)
and EmbeddGAN ($0.469$) are competitive, while SN-GAN ($0.328$) underperforms.
And on the CelebA dataset, WGAN-GP achieves the highest recall ($0.495$), with
SN-GAN ($0.471$) and DCGAN ($0.376$) trailing, and EmbeddGAN ($0.352$) with the
lowest.

\subsection{Qualitative Comparison}

\begin{figure*}
	\centering
	\includegraphics[width=.8\linewidth]{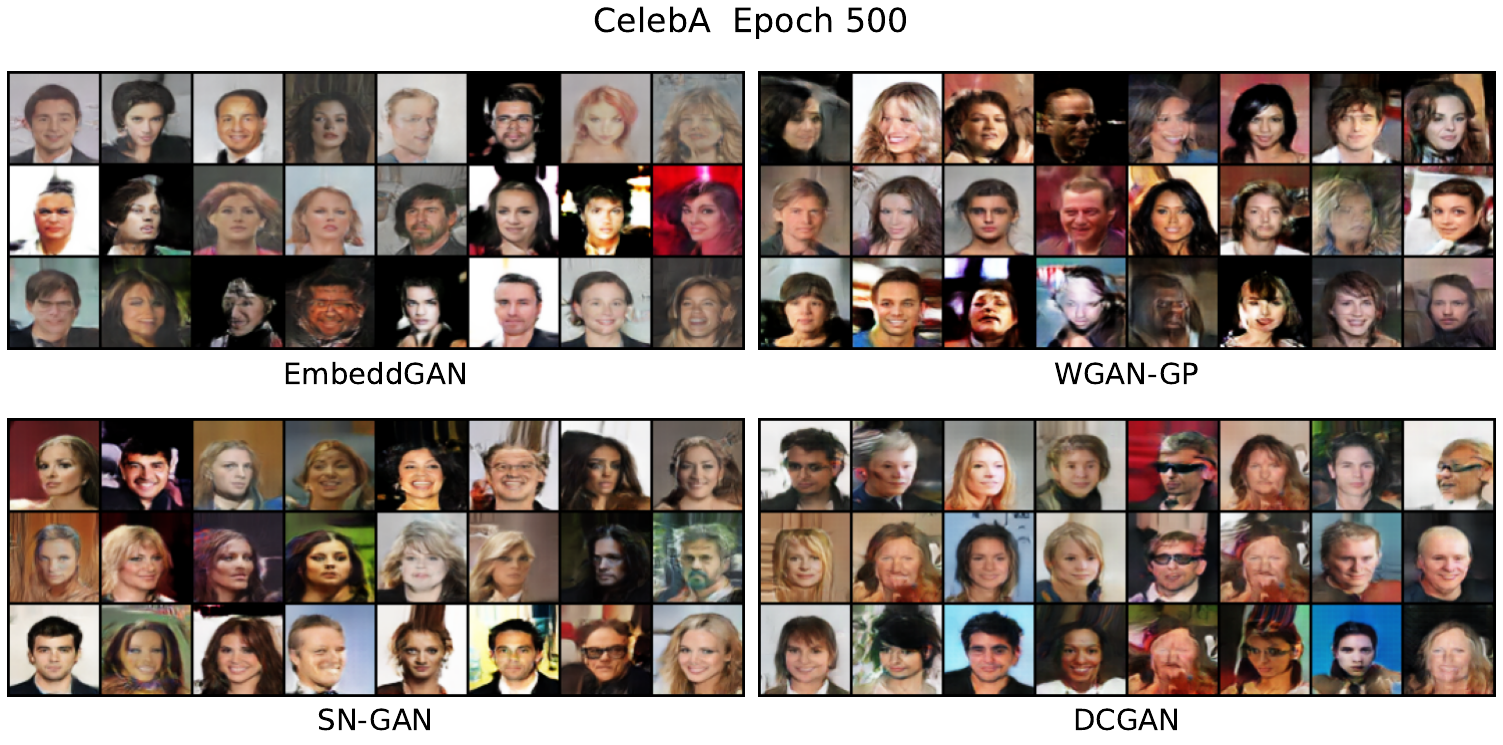}
	\caption{Generated samples from EmbeddGAN, WGAN-GP, SN-GAN and DCGAN on
		CelebA at epoch 500.}
	\label{fig:celeba_comparison}
\end{figure*}

On CelebA, Fig.~\ref{fig:celeba_comparison} shows randomly sampled outputs from
all four models at epoch 500. EmbeddGAN produces recognizable, structurally
coherent faces with consistent facial features and diversity in attributes such
as hair color and expression, demonstrating that the gCor-based loss
successfully drives the generator toward the real data distribution. A subset
of samples shows desaturation relative to WGAN-GP and SN-GAN, we discuss
possible causes of this in section~\ref{sec:limitations}. WGAN-GP and SN-GAN
produce good outputs consistent with their strong FID scores, while DCGAN shows
moderate sample-to-sample variation in color and sharpness.
\begin{figure*}
	\centering
	\includegraphics[width=.8\linewidth]{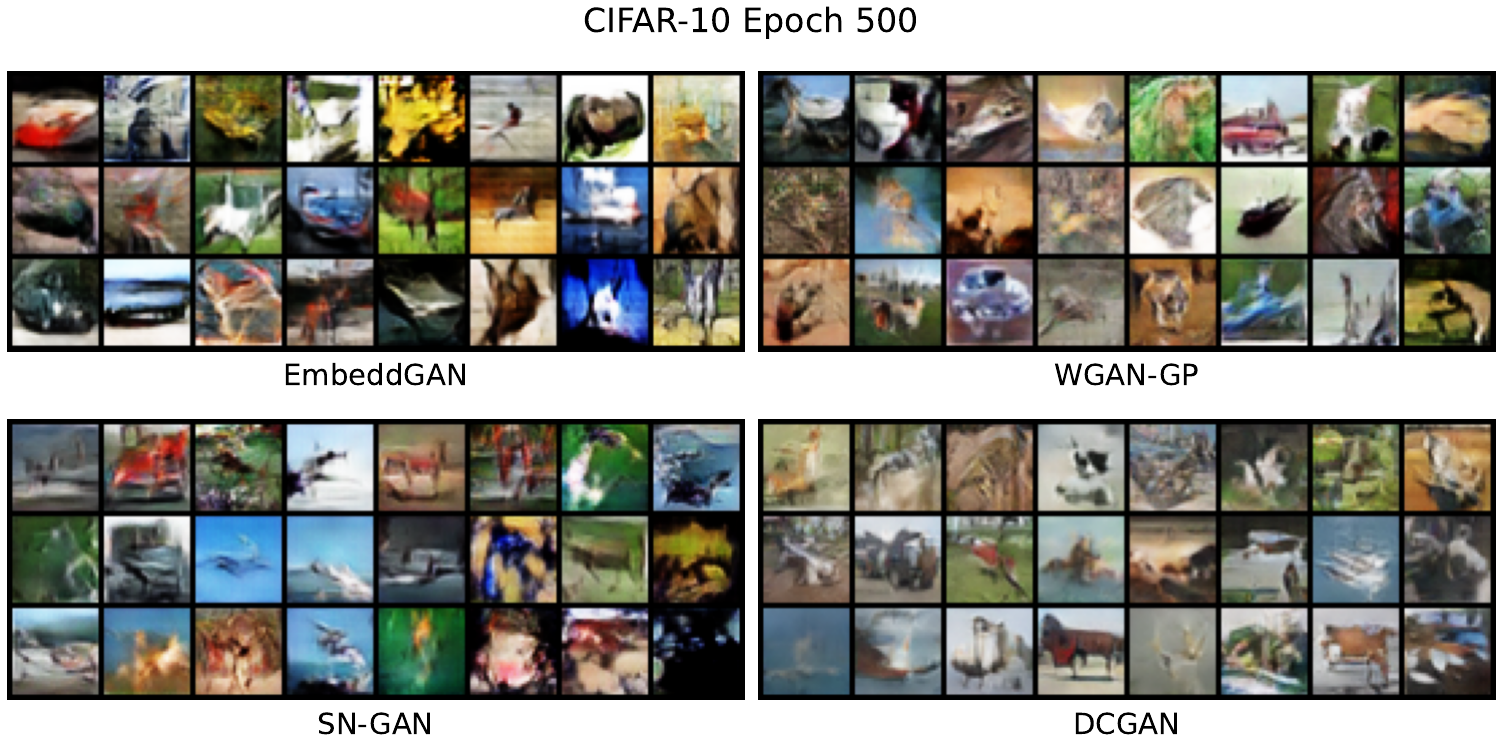}
	\caption{Generated samples from EmbeddGAN, WGAN-GP, SN-GAN and DCGAN on
		CIFAR-10 at epoch 500.}
	\label{fig:cifar10_comparison}
\end{figure*}
On the CIFAR-10 comparison shown in Fig.~\ref{fig:cifar10_comparison}, EmbeddGAN
performs well against all three baselines, showing good diversity and quality
in the generated samples.

\subsection{Training Stability}

\begin{figure*}[]
	\centering
	\subfloat[MNIST]{\includegraphics[width=0.33\textwidth]{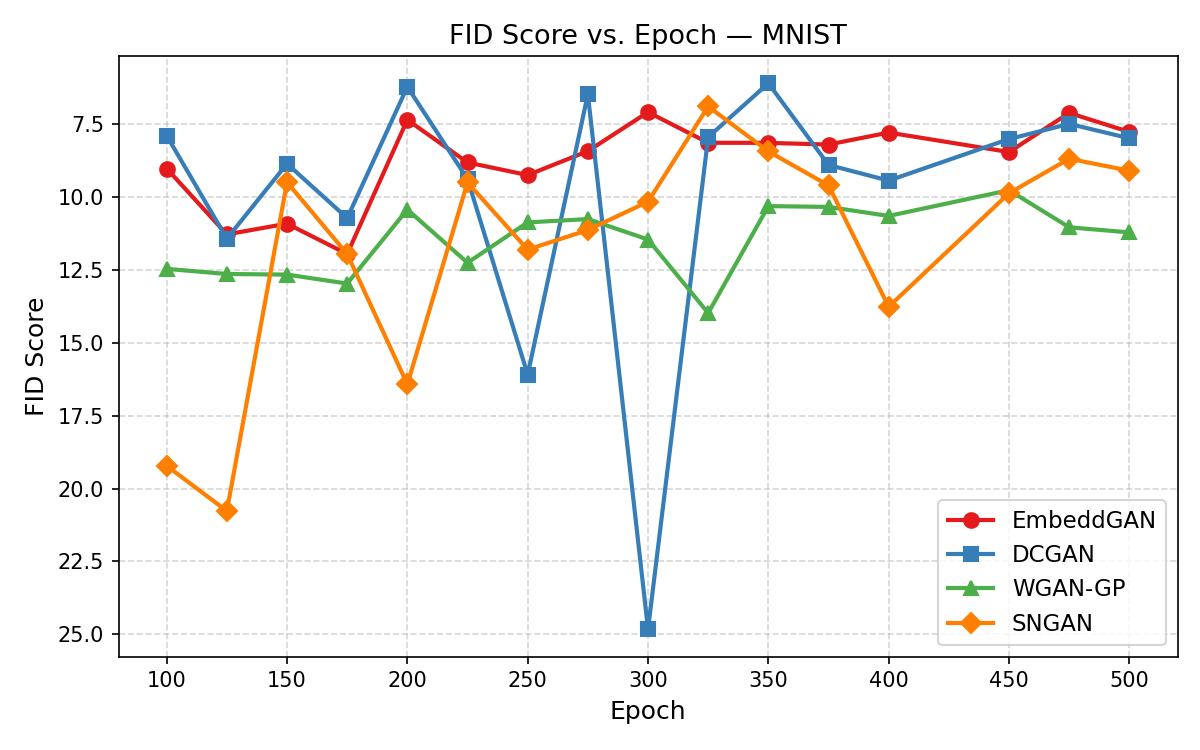}}
	\hfill
	\subfloat[CIFAR-10]{\includegraphics[width=0.33\textwidth]{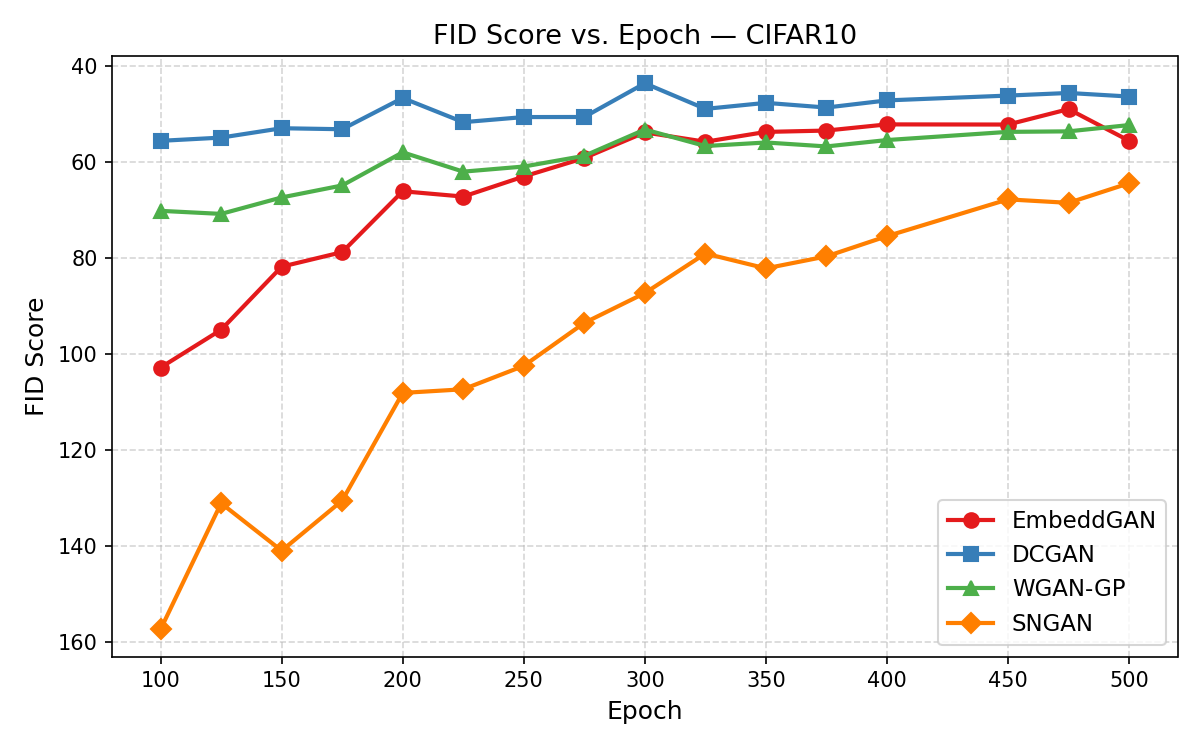}}
	\hfill
	\subfloat[CelebA]{\includegraphics[width=0.33\textwidth]{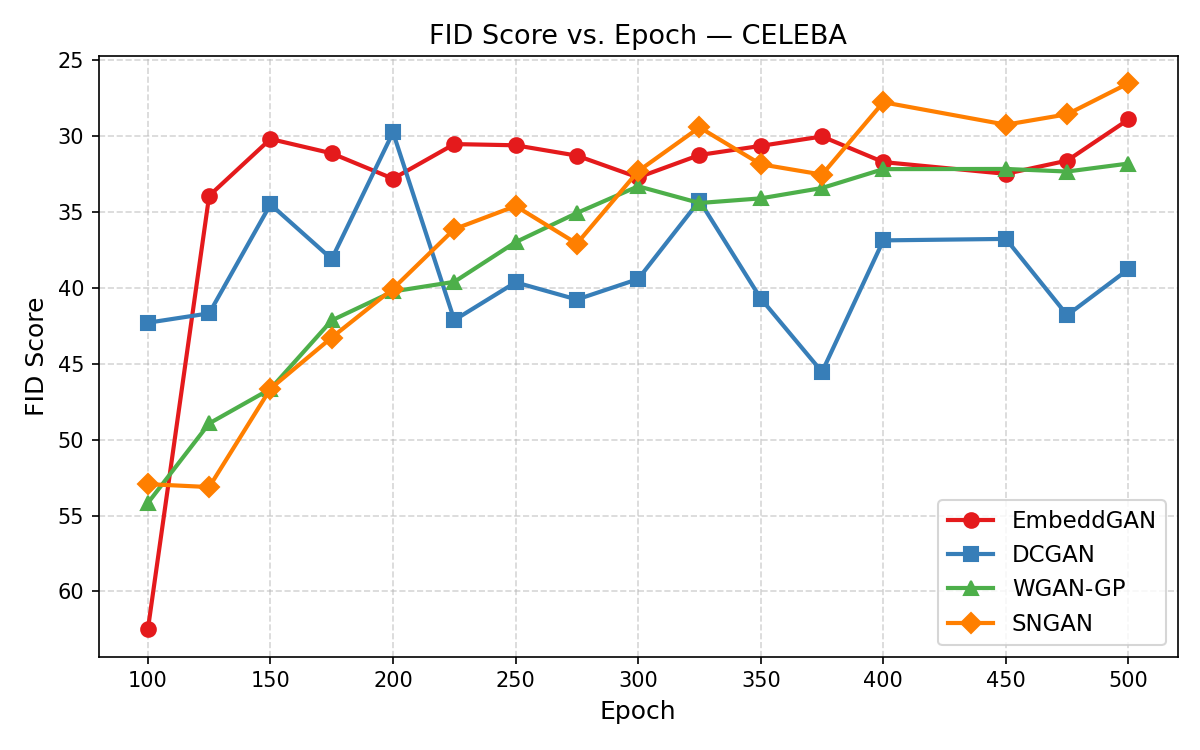}}
	\caption{FID score vs.\ training epoch for EmbeddGAN, DCGAN, WGAN-GP, and SN-GAN
		on MNIST, CIFAR-10, and CelebA. Computed over a sample size of $5000$. Lower is better. EmbeddGAN maintains
		consistent stability on MNIST across all epochs and matches WGAN-GP on
		CIFAR-10 by epoch 300. DCGAN exhibits substantial oscillation on CelebA
		and occasional instability on MNIST.}
	\label{fig:trajectory}
\end{figure*}

Fig.~\ref{fig:trajectory} shows FID score as a function of training epoch for
each model on all three datasets. To manage the computational overhead of dense
evaluation across training, FID is estimated using $5,000$ generated samples at
each checkpoint rather than the standard $50,000$. On MNIST, EmbeddGAN reaches
near-optimal FID by epoch 100 and remains consistently stable through epoch
500, never exhibiting a degraded checkpoint. DCGAN achieves a comparable final
FID but shows occasional instability --- most notably a spike to FID $25$ at
epoch 300 --- before recovering by epoch 325. On CIFAR-10, EmbeddGAN starts as
the second-weakest model but improves steadily throughout training, matching
WGAN-GP by epoch 300 and maintaining parity through epoch 500, while DCGAN
leads throughout. On CelebA, EmbeddGAN undergoes rapid improvement in the first
150 epochs, reducing FID from $62$ to approximately $30$, before plateauing for
the remainder of training. DCGAN exhibits substantial oscillation on CelebA
throughout training, with epoch-to-epoch variance of up to 15 FID points.
WGAN-GP converges steadily but reaches EmbeddGAN's plateau only after epoch
400.

\subsection{Loss Curve}

\begin{figure*}[]
	\centering
	\subfloat[MNIST]{\includegraphics[width=0.33\textwidth]{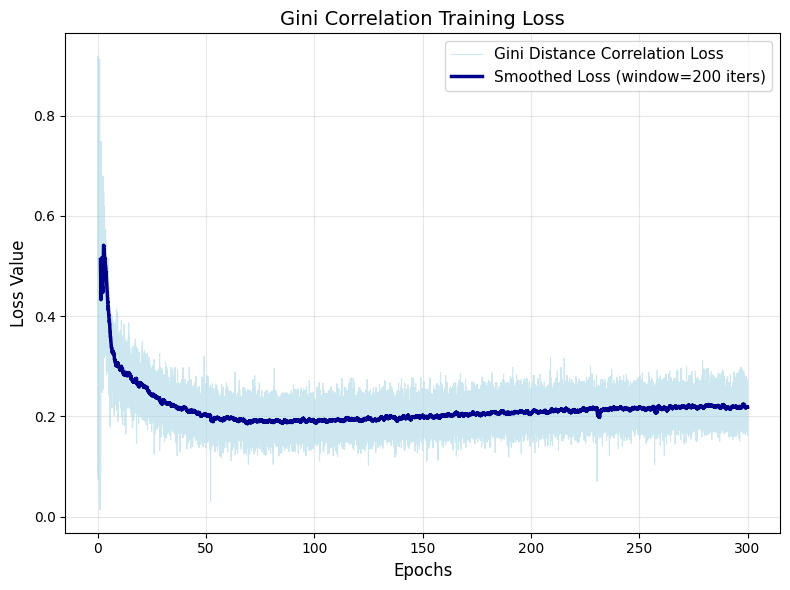}}
	\hfill
	\subfloat[CIFAR-10]{\includegraphics[width=0.33\textwidth]{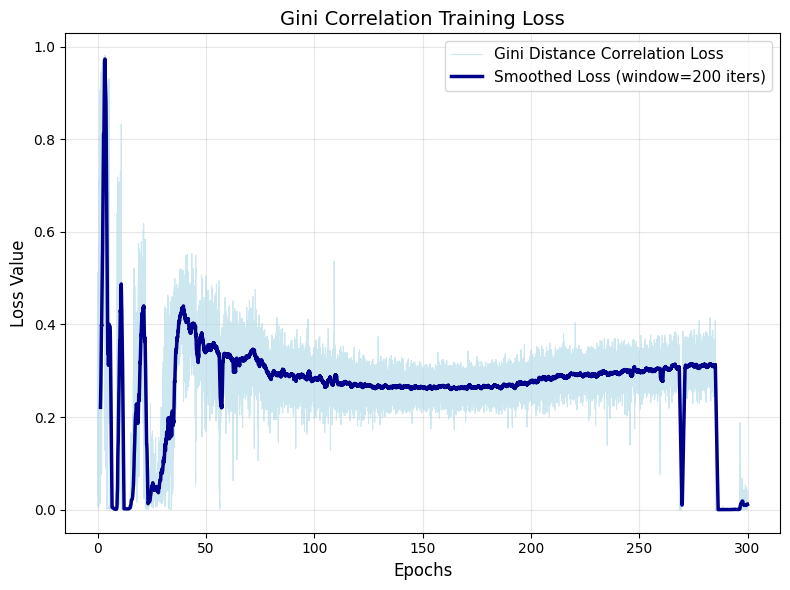}}
	\hfill
	\subfloat[CelebA]{\includegraphics[width=0.33\linewidth]{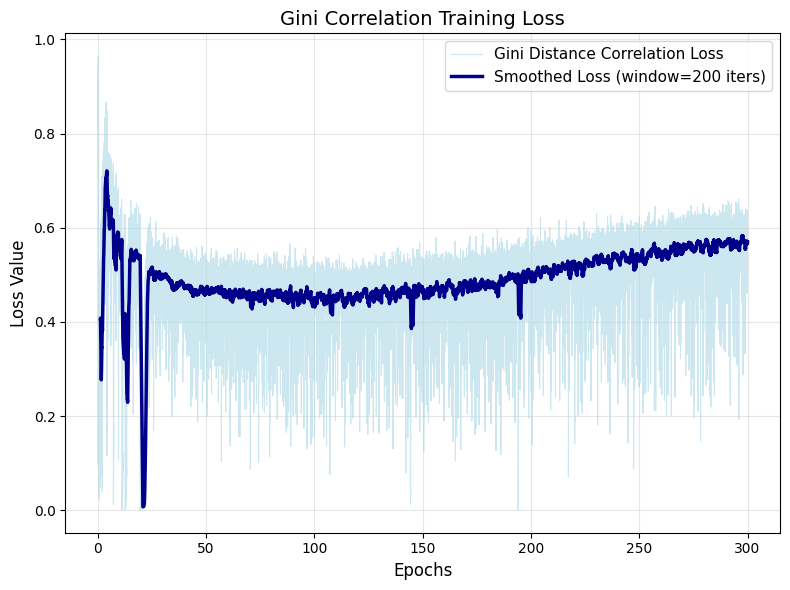}}
	\caption{Gini distance correlation loss curve for EmbeddGAN on MNIST, CIFAR-10, and CelebA.
		MNIST converges quickly to a stable plateau near $0.2$. CIFAR-10 shows high early volatility
		before settling around $0.25$--$0.30$, with a brief collapse near epoch $270$. CelebA
		stabilizes around $0.45$--$0.50$ but drifts upward through training, indicating the
		embedding network maintains a persistent advantage over the generator on the more complex dataset.}
	\label{fig:embeddgan_loss}
\end{figure*}

Fig.~\ref{fig:embeddgan_loss} shows the Gini distance correlation loss for the
EmbeddGAN generator on all three datasets with L2 normalization and R1
regularization. In all cases, the loss spikes early as the embedding network
learns to separate real from generated samples, then falls as the generator
catches up. On MNIST, the loss drops from around $0.5$ to a stable plateau near
$0.2$ by epoch $50$ and stays there. On CIFAR-10, training is much noisier. The
loss swings between near-zero and $0.45$ for the first $70$ epochs before
settling around $0.25$--$0.30$. There is also a brief collapse near epoch $270$
where the loss drops to near zero before recovering. On CelebA, the loss
settles around $0.45$--$0.50$ by epoch $25$ but slowly drifts upward to around
$0.55$--$0.60$ by epoch $300$, suggesting the embedding network continues to
pull ahead on this more complex dataset, with the generator unable to fully
close the gap.

\subsection{Ablation Studies}

\label{sec:ablation_study}

\begin{figure}[]
	\centering
	\includegraphics[width=\linewidth]{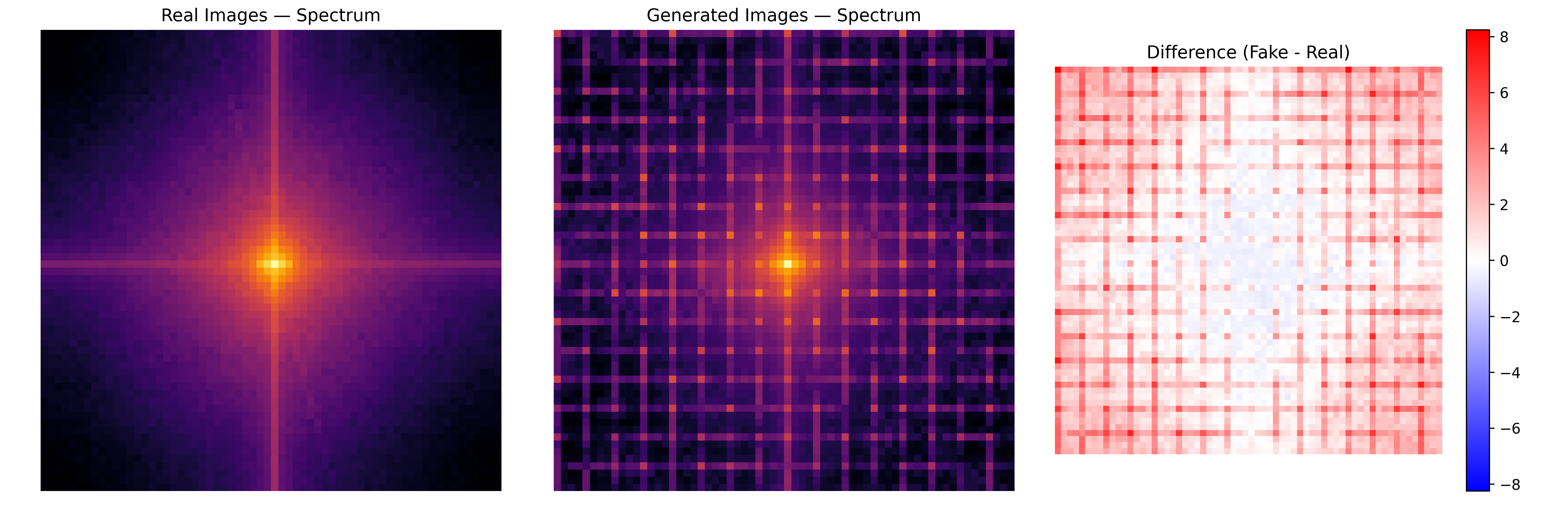}

	\caption{Average 2D Fourier power spectrum of generated and real
		CelebA images during unregularized EmbeddGAN training. The generated spectrum
		exhibits a periodic grid of bright peaks at evenly spaced
		frequencies, characteristic of transposed convolution
		checkerboard aliasing.}
	\label{fig:fourier}
\end{figure}

\begin{figure}[]
	\centering
	\includegraphics[width=\linewidth]{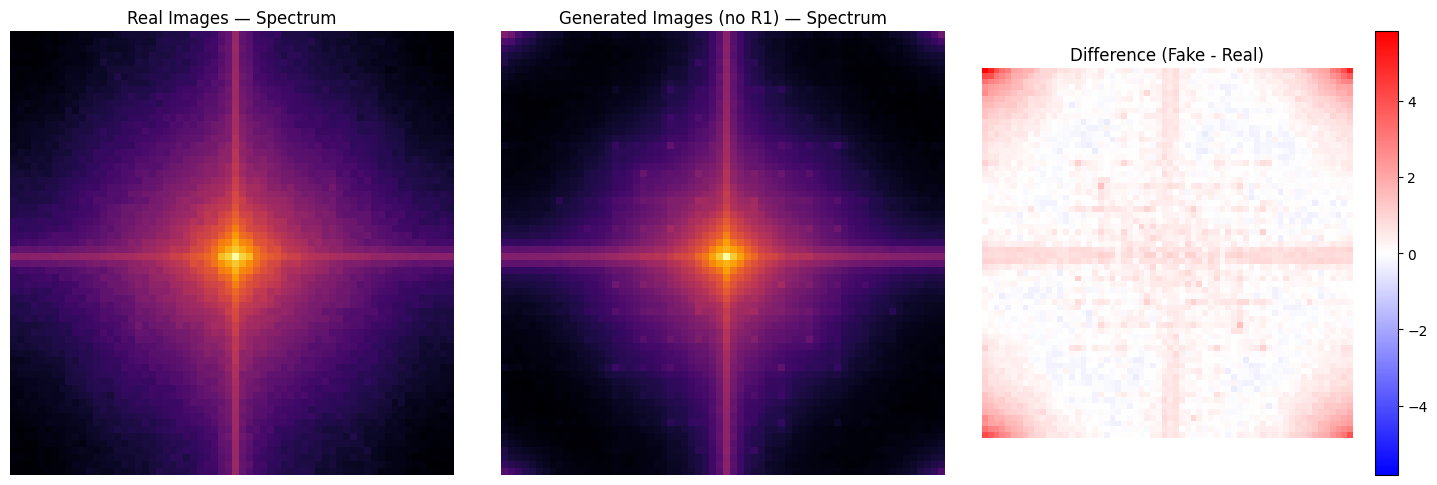}

	\caption{Adding L2 normalization alone eliminates the periodic grid of peaks,
		confirming that constraining embeddings to the unit hypersphere closes the
		aliasing shortcut. The generated spectrum remains noisy overall, however,
		with the difference map showing broadly distributed excess high-frequency
		energy, indicating that L2 suppresses the structural artifact but cannot
		align the generator's output with the real spectral distribution.}
	\label{fig:fourier_with_l2_no_r1}
\end{figure}

This section provides quantitative evidence for three design claims: the
necessity of joint L2/R1 regularization, the sensitivity of gCor to embedding
dimensions, and the importance of balanced update dynamics.

\begin{figure}[]
	\centering
	\includegraphics[width=\linewidth]{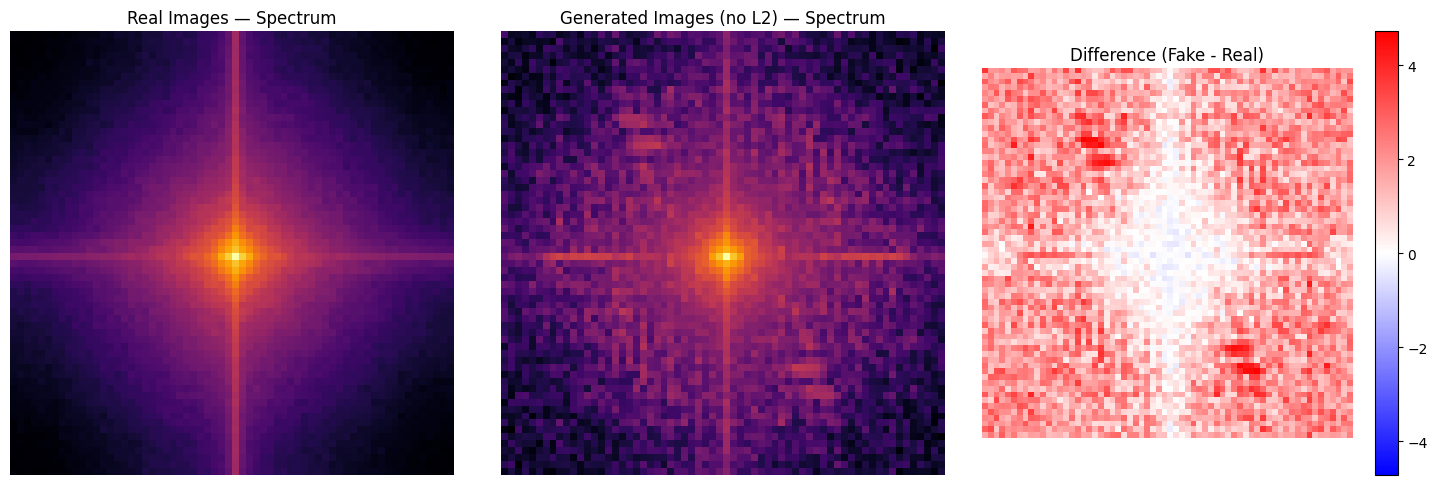}

	\caption{R1 regularization alone also eliminates the periodic grid, but via
		a different mechanism: penalizing gradients on real images removes the
		training incentive to exploit aliasing rather than constraining the
		embedding space. The spectrum remains unstructured, with excess
		high-frequency energy distributed throughout the difference map,
		mirroring the residual noise seen with L2 alone and indicating that
		neither component is sufficient on its own.}
	\label{fig:fourier_no_l2}
\end{figure}

\begin{figure}[]
	\centering
	\includegraphics[width=\linewidth]{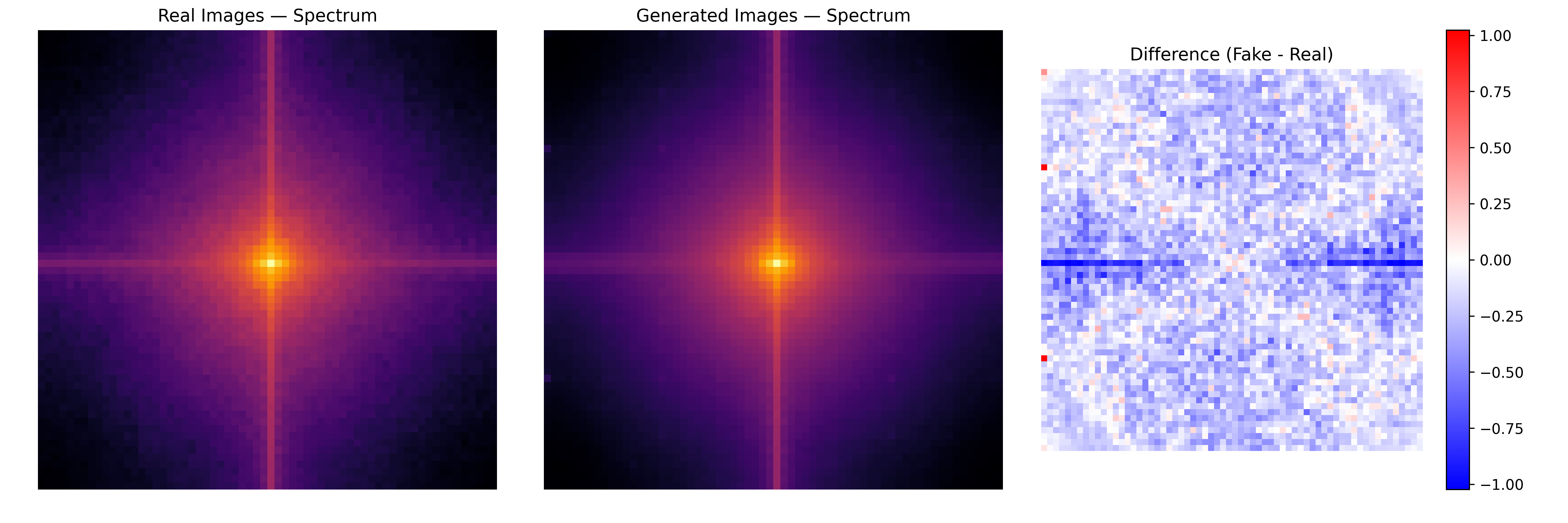}

	\caption{Average 2D Fourier power spectrum of generated and real
		CelebA images after the introduction of L2 normalization and
		R1 regularization in EmbeddGAN at epoch 300. The periodic peaks observed in
		the generated spectrum are substantially reduced, indicating
		that the tiling artifacts have been mitigated.}
	\label{fig:fourier_with_l2_r1}
\end{figure}

We perform our ablation studies using the CelebA dataset, as it is both the
most generally complex and produces reasonably good images, as seen in
Fig.~\ref{fig:celeba_comparison}. To verify that L2 normalization and R1
regularization each address the distinct failure modes described in
Section~\ref{sec:regularization}, we computed the average 2D Fourier power
spectrum over generated and real CelebA images across four regularization
conditions. Fourier power spectrum analysis provides a frequency-domain view of
the generated image statistics, allowing us to distinguish structured aliasing
artifacts, which appear as periodic peaks, from the broadband spectral noise
associated with shortcut exploitation. Fig.~\ref{fig:fourier} shows that
without any regularization, the generated spectrum contains a periodic grid of
bright peaks at evenly-spaced frequencies, while the real image spectrum
exhibits smooth radial decay consistent with natural image statistics. This
pattern is characteristic of transposed convolution checkerboard aliasing,
arising from uneven kernel overlap across the generator's upsampling stages,
and confirms the tiling artifact as an \emph{architectural} property of the
generator rather than a consequence of the training objective.
Fig.~\ref{fig:fourier_with_l2_no_r1} shows that adding L2 normalization alone
significantly lessens the periodic grid by constraining the embedding space to
the unit hypersphere, but the generated spectrum remains globally noisy, with
the difference map showing broadly distributed excess high-frequency energy.
Fig.~\ref{fig:fourier_no_l2} shows that R1 regularization alone also
significantly lessens the periodic grid, this time by removing the gradient
incentive to exploit aliasing, yet the spectrum remains similarly unstructured.
That both components independently suppress the aliasing shortcut through
different mechanisms, yet neither resolves the spectral mismatch alone,
suggests the two play complementary roles. Fig.~\ref{fig:fourier_with_l2_r1}
confirms this: combining L2 normalization with R1 brings the generated spectrum
closer to the real images, with the difference map reduced to near zero.

\subsubsection{Joint Necessity of L2 Normalization and R1 Regularization}

\begin{figure*}[]
	\centering

	\subfloat[No L2, no R1 (epoch 300)\label{fig:no_l2_or_r1}]{%
		\includegraphics[width=0.4\textwidth]{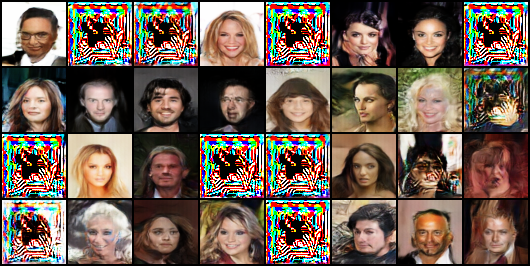}}
	\ \ \ \
	\subfloat[L2 only, no R1 (epoch 300)\label{fig:no_r1}]{%
		\includegraphics[width=0.4\textwidth]{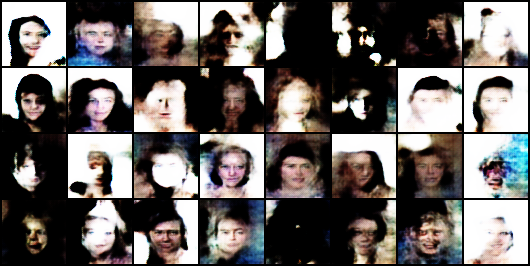}}

	\subfloat[R1 only, no L2 (epoch 300)\label{fig:no_l2}]{%
		\includegraphics[width=0.4\textwidth]{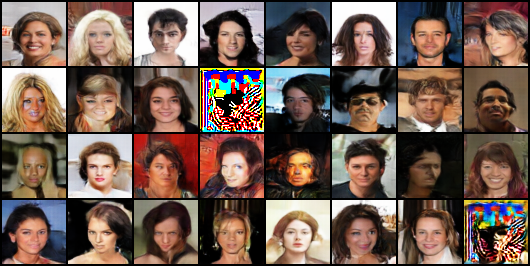}}
	\ \ \ \
	\subfloat[L2 + R1 (epoch 300)\label{fig:with_l2_r1}]{%
		\includegraphics[width=0.4\textwidth]{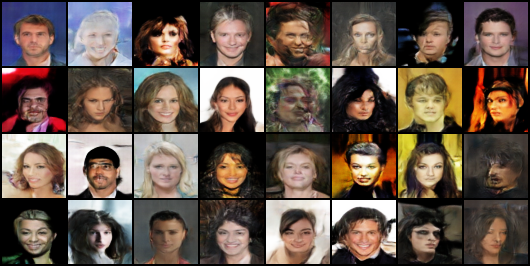}}

	\caption{Generated CelebA samples showing the development of EmbeddGAN's
		regularization strategy. (a)~Without either component, the generator produces
		colorful tiling artifacts. (b)~L2 alone resolves the magnitude shortcut,
		but the embedding network finds high-contrast regions as the next shortcut,
		causing contrast collapse. (c)~R1 alone does not prevent the magnitude shortcut
		when embeddings are unbounded.  (d)~With both L2 and R1, both escape routes are
		closed, and the generator produces clean, diverse faces.}
	\label{fig:ablation}
\end{figure*}

With the understanding that the exploitation visible in Fig.~\ref{fig:fourier}
is limited with the combination of both L2 normalization and R1 regularization,
we examine the effect of the L2 normalization and R1 gradient regularization
independently. We train four total models of EmbeddGAN, without L2/R1, with L2
without R1, without L2 with R1, and with both L2 and R1, for 300 epochs. We
examine both qualitatively and quantitatively the results for comparison.
Initially, we remove both L2 normalization and R1 gradient regularization,
resulting in a colorful tiling indicating mode collapse.
Fig.~\ref{fig:ablation}(a) shows an example of this issue. We then add L2
normalization back to determine if it alone can fix the tiling issue.
Fig.~\ref{fig:ablation}(b) shows these results, with high contrast images
dominating, implying a shortcut has been found by the generator. Based on the
Fourier power spectrums we previously examined, L2 alone closes the magnitude
shortcut, but in doing so discovers another shortcut (the use of high
contrast). Applying only R1 gradient regularization
(Fig.~\ref{fig:ablation}(c)) shows a new, less regular instability occurs, but
the faces that are shown are clearly sharper. Additionally, the high contrast
shortcut discovered by L2 normalization is seemingly closed, implying that, to
fix instability in this model, both techniques are required. Combining both L2
normalization and R1 gradient regularization, as shown in Fig.~
\ref{fig:ablation}(d), results in consistent, sharp faces.

Table~\ref{tab:ablation_fid_scores} quantifies these outcomes on CelebA FID at
epoch 300. Removing both components raises FID from $29.29$ to $70.22$. The
most striking result is L2 alone, which produces the worst FID of $179.21$,
substantially higher than having no regularization at all. This is the
counterintuitive consequence described above: closing the magnitude shortcut
forces the embedding network toward the contrast shortcut, and the resulting
collapse is more damaging than the unconstrained multi-shortcut behavior seen
without either component. R1 alone ($54.29$) provides meaningful improvement,
but the remaining magnitude shortcut prevents the embedding network from
measuring distributional overlap reliably. Together, L2 and R1 are
complementary rather than redundant: each targets a distinct failure mode the
other cannot address.

\begin{table}[]
	\centering
	\caption{Ablation study of EmbeddGAN on CelebA at epoch 300 over $50000$ samples. Removing
		either regularization component leads to a substantial increase in FID;
		L2 normalization alone produces the worst result, as it forces the
		embedding network toward the contrast shortcut.}
	\begin{tabular}{lc}
		\toprule\toprule
		Variant                  & CelebA FID $\downarrow$ \\
		\midrule
		Full EmbeddGAN (L2 + R1) & $29.29$                 \\
		R1 only, no L2           & $54.29$                 \\
		No L2, no R1             & $70.22$                 \\
		L2 only, no R1           & $179.21$                \\
		\bottomrule
	\end{tabular}
	\label{tab:ablation_fid_scores}
\end{table}

\subsubsection{Ablation of Embedding Dimension}

\begin{table}[]
	\centering
	\caption{Ablation study of EmbeddGAN with varying embedding dimensions on CelebA at epoch 500 over $50000$ samples.
		Comparison $n=100$ in bold.
	}
	\begin{tabular}{lc}
		\toprule\toprule
		Embedding Dims & CelebA FID $\downarrow$ \\
		\midrule
		1              & $411.39$                \\
		3              & $28.31$                 \\
		5              & $27.09$                 \\
		10             & $29.25$                 \\
		50             & $39.18$                 \\
		\textbf{100}   & $\mathbf{32.09}$        \\
		1000           & $32.29$                 \\
		\bottomrule
	\end{tabular}
	\label{tab:n_dims_ablation_fid_scores}
\end{table}

Table~\ref{tab:n_dims_ablation_fid_scores} shows the FID scores for different
embedding dimensions. A single embedding dimension ($n=1$) is insufficient,
yielding a FID of $411.39$ on the CelebA dataset. The distributional signal
collapsed to a scalar is too poor a proxy for the real/generated gap.
Performance improves sharply at $n=3$ (FID $28.31$) and peaks at $n=5$ (FID
$27.09$), indicating that even a very low-dimensional embedding space is
sufficient to capture the distributional structure needed for gCor. Beyond this
spot, FID degrades modestly: $n=10$ scores $29.25$, $n=50$ scores $39.18$, and
$n=1000$ scores $32.29$. This U-shaped trend could suggest that very high
dimensionality dilutes the gCor signal with noise. These results support using
$n\geq 3$ in practice, and motivate our default choice of $n=100$ as a
conservative choice well above the minimum threshold.

\begin{figure}
	\centering
	\includegraphics[width=0.6\linewidth]{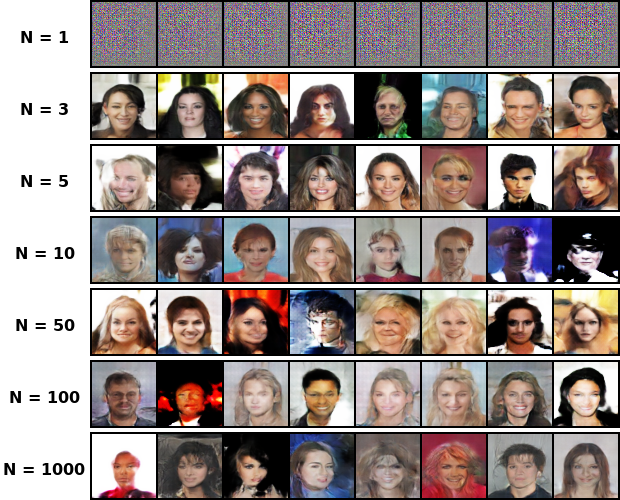}
	\caption{Qualitative comparison of generated CelebA samples from EmbeddGAN with
		varying embedding dimensions after 500 epochs.}

	\label{fig:qualitative_ndims_comparison}
\end{figure}

Fig.~\ref{fig:qualitative_ndims_comparison} provides a qualitative comparison
of generated CelebA samples from EmbeddGAN with varying embedding dimensions.
Samples are generated randomly. As suggested by the FID scores, with $n=1$, the
generator fails to learn meaningful structure producing random noise. At
$n\geq3$ we can see dramatic improvement with recognizable faces being
generated, supporting the lower FID values. Importantly, while $n=3$, $n=5$,
and $n=10$ produced better FID values, their qualitative results from
Fig.~\ref{fig:qualitative_ndims_comparison} are not noticeably improved
compared to our reference $n=100$.

\subsubsection{Ablation of Embedding Steps per Generator Update}

\begin{table}[]
	\centering
	\caption{Ablation study of EmbeddGAN with varying embedding net steps per generator update on CelebA at epoch 300 over $50000$ samples. }
	\begin{tabular}{lc}
		\toprule\toprule
		k\_steps & CelebA FID $\downarrow$ \\
		\midrule
		1        & $29.29$                 \\
		3        & $43.15$                 \\
		5        & $47.54$                 \\
		10       & $131.36$                \\
		\bottomrule
	\end{tabular}
	\label{tab:k_steps_ablation_fid_scores}
\end{table}

Table~\ref{tab:k_steps_ablation_fid_scores} shows the FID scores for different
numbers of embedding steps per generator update. Increasing the number of
embedding network update steps ($k$) per generator update consistently degrades
performance. At $k=3$, FID is $43.15$; at $k=5$ it rises to $47.45$; and at
$k=10$ it climbs to $131.36$. This monotonic degradation suggests that
over-training the embedding network relative to the generator creates an
increasingly adversarial dynamic, where the embedding network pulls so far
ahead that the gCor signal become uninformative or unstable for the generator.
This mirrors the well-known instability observed in WGAN-GP when the
critic-to-generator step ratio is pushed too high. The results motivate our
default of $k=1$, which maintains a balanced update dynamic and achieves the
best performance in our experiments.

\begin{figure}[h]
	\centering
	\includegraphics[width=0.7\linewidth]{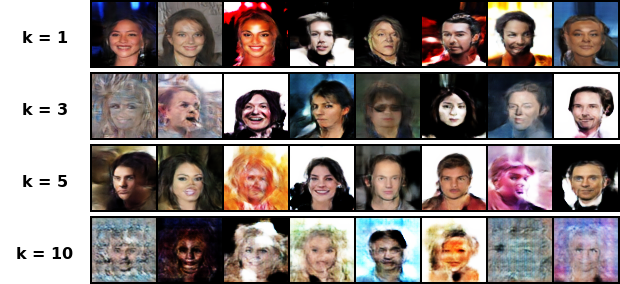}
	\caption{Qualitative comparison of generated CelebA samples from EmbeddGAN with varying
		numbers of embedding steps per generator update after 300 epochs.}
	\label{fig:qualitative_ksteps_comparison}
\end{figure}

As shown in Fig.~\ref{fig:qualitative_ksteps_comparison}, with $k=1$, the
generator produces recognizable, diverse faces. At $k=3$, the quality degrades
with more blurring and less diversity. At $k=5$, the samples seem to continue
to degrade with some recognizable faces. Setting $k=10$ severely degrades the
quality, the generator fails to learn meaningful structure, producing mostly
noise and blurry faces. This qualitative degradation is consistent with the FID
scores and supports the conclusion that a balanced update dynamic with $k=1$ is
crucial for stable and effective training in EmbeddGAN.

\subsection{Visualizing the Embedding Space}

\begin{figure*}[h]
	\centering
	\includegraphics[width=0.9\textwidth]{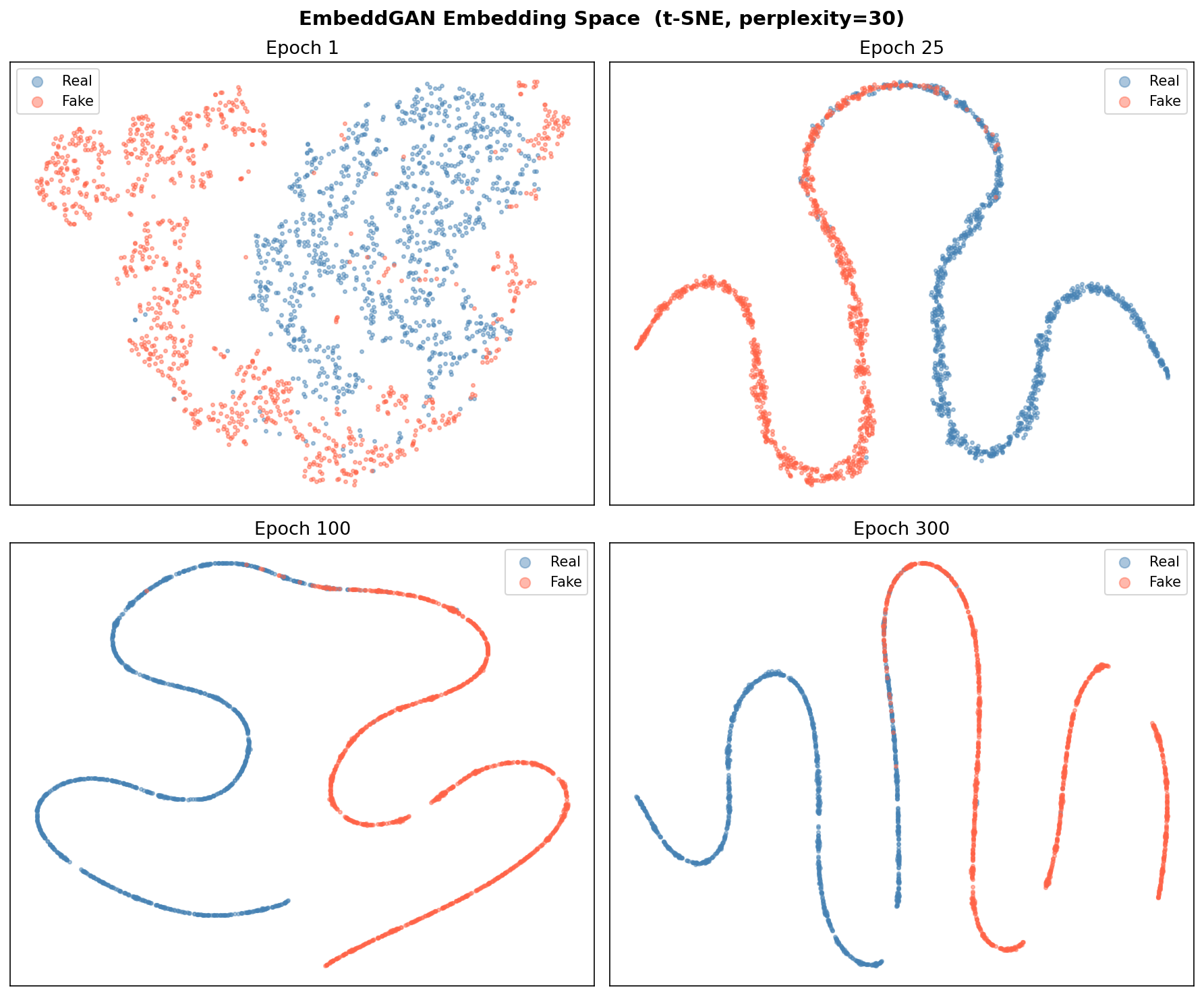}
	\caption{Visualization of the EmbeddGAN embedding space on CelebA using t-SNE at
		epochs 1, 25, 100, and 300. At epoch 1, real and generated embeddings are
		intermixed with no discernible structure. By epoch 25, the two
		distributions have separated into distinct curved manifolds in the projected
		space, and this separation is maintained at epoch 100 and 300, providing a
		persistent distributional signal for gCor to measure.}
	\label{fig:embedding_space}
\end{figure*}

One of the key advantages of EmbeddGAN's embedding-based framework is the
ability to directly visualize the learned embedding space and the separation
between real and generated distributions. Fig.~\ref{fig:embedding_space}
visualizes the embedding space learned by the embedding network on CelebA using
t-SNE~\cite{vandermaatenVisualizingDataUsing2008}. At epoch 1, real and
generated embeddings are intermixed with no discernible structure, reflecting
the random initialization of the embedding network before any meaningful
training signal has propagated. By epoch 25, the two distributions have
separated into distinct curved manifolds in the projected space, and by epoch
100 this separation is maintained. This separation is the direct objective of
the embedding network: by maximizing gCor, the embedding network is
incentivized to find projections that make the real and generated distributions
statistically distinguishable. The curved manifold structure, rather than tight
point clusters, suggests a distributed representation capturing diversity
within each distribution rather than collapsing to a single mode. This rapid
separation by epoch 25 is consistent with EmbeddGAN's fast initial FID
improvement on CelebA observed in Fig.~\ref{fig:trajectory}.

\subsection{Limitations}
\label{sec:limitations}

EmbeddGAN exhibits a mean-seeking failure mode in late training on CelebA,
where generated images progressively lose color saturation and converge toward
a gray average. We hypothesize that R1 regularization, applied at every
training step over hundreds of epochs, cumulatively compresses the sensitivity
of the embedding network, producing an increasingly tight embedding cluster for
real images. As this cluster tightens, the gradient signal available to the
generator diminishes, and the generator drifts toward the cluster mean rather
than the full distribution. Reducing \texttt{R1\_GAMMA} from $10.0$ to $0.5$
reduced its severity, suggesting that the aggressiveness of regularization is
the primary driver. Lazy regularization, computing the R1 penalty every $N$
steps rather than every step, as used in
StyleGAN2~\cite{karrasAnalyzingImprovingImage2020}, remains a candidate fix for
this failure mode.

\section{Conclusion}
\label{sec:conclusion}

This paper presented EmbeddGAN, a novel GAN training framework that replaces
the traditional discriminator with an embedding network trained via Gini
distance correlation (gCor). By recasting the adversarial objective as a
minimax problem over a learned embedding space, where the embedding network
maximizes gCor and the generator minimizes it, EmbeddGAN grounds GAN training
in a statistically formal measure of distributional dependence with a clear
convergence condition: gCor equals zero if and only if the real and generated
distributions are indistinguishable in the learned embedding space. Experiments
on MNIST, CIFAR-10, and CelebA demonstrated that EmbeddGAN achieves competitive
FID scores against DCGAN, WGAN-GP, and SN-GAN across all three datasets, while
exhibiting notably more stable training dynamics than DCGAN on CelebA. Ablation
studies revealed that L2 normalization and R1 gradient regularization are
jointly necessary to prevent distinct failure modes, magnitude exploitation,
and high-contrast shortcuts that arise when the embedding network is
unconstrained. These results suggest that binary classification is not the only
viable adversarial signal; dependence-based objectives in learned embedding
spaces offer a promising alternative, though their success depends critically
on regularizing the embedding network to prevent shortcut solutions.
%Bibliography
\bibliographystyle{IEEEtran}
\bibliography{refs}

\end{document}